\documentclass[10pt,twocolumn,letterpaper]{article}

\usepackage{cvpr}
\usepackage{times}
\usepackage{epsfig}
\usepackage{graphicx}
\usepackage{amsmath}
\usepackage{amssymb}

\usepackage{textcomp} 
\usepackage{gensymb} 
\usepackage{array} 
\usepackage{booktabs} 
\usepackage{ragged2e} 
\usepackage{xcolor} 
\usepackage{subcaption}
\usepackage{cite}
\usepackage[utf8]{inputenc}

\usepackage[pagebackref=true,breaklinks=true,letterpaper=true,colorlinks,bookmarks=false]{hyperref}

\cvprfinalcopy 

\def\cvprPaperID{****} 

\ifcvprfinal\fi
\begin{document}

\title{Solo or Ensemble?\\Choosing a CNN Architecture for Melanoma Classification}

\author{Fábio Perez\textsuperscript{1} ~~~~~~ Sandra Avila\textsuperscript{2} ~~~~~~ Eduardo Valle\textsuperscript{1} \\
{\tt\small \{fperez, dovalle\}@dca.fee.unicamp.br} ~~~~~~~ {\tt\small sandra@ic.unicamp.br} \\ \\
\textsuperscript{1}School of Electrical and Computing Engineering (FEEC)~~~\textsuperscript{2}Institute of Computing (IC) \\
RECOD Lab, University of Campinas (UNICAMP), Brazil \\ \\
}

\maketitle
\thispagestyle{empty}

\begin{abstract}
Convolutional neural networks (CNNs) deliver exceptional results for computer vision, including medical image analysis. With the growing number of available architectures, picking one over another is far from obvious. Existing art suggests that, when performing transfer learning, the performance of CNN architectures on ImageNet correlates strongly with their performance on target tasks. We evaluate that claim for melanoma classification, over 9 CNNs architectures, in 5 sets of splits created on the ISIC Challenge 2017 dataset, and 3 repeated measures, resulting in 135 models. The correlations we found were, to begin with, much smaller than those reported by existing art, and disappeared altogether when we considered only the top-performing networks: uncontrolled nuisances (i.e., splits and randomness) overcome any of the analyzed factors. Whenever possible, the best approach for melanoma classification is still to create ensembles of multiple models. We compared two choices for selecting which models to ensemble: picking them at random (among a pool of high-quality ones) vs. using the validation set to determine which ones to pick first. For small ensembles, we found a slight advantage on the second approach but found that random choice was also competitive. Although our aim in this paper was not to maximize performance, we easily reached AUCs comparable to the first place on the ISIC Challenge 2017.
\end{abstract}


\section{Introduction}
\label{sec:introduction}

Deep learning has achieved impressive results in skin lesion analysis, including lesion segmentation, lesion classification, and medical attribute detection. Since 2015, convolutional neural networks (CNNs) are the state of the art for melanoma classification~\cite{data_depth_design}. The standard procedure for training melanoma classification models is to fine tune an ImageNet pre-trained CNN for a melanoma dataset~\cite{transfer_afonso}. However, with the crescent number of CNN architectures, choosing the one to employ is increasingly difficult for researchers.

The ISIC Challenge~\cite{isic_challenge_2016, isic_challenge_2017, isic_challenge_2018}, the largest skin lesion analysis competition, illustrates such increase in the number of available CNN architectures. In the first two editions of the Challenge, the most successful submissions featured mostly ResNet and Inception architectures. In contrast, the most recent, third edition (in 2018) showcased a much wider range of architectures, including also Dual Path Networks, Squeeze-and-Excitation Networks, PNASNet, among others. (We discuss those architectures in Section~\ref{sec:methodology}).

Designing the right CNN architecture for skin lesion analysis (or, in fact, for any image classification task) is far from obvious. Since most tasks tend to reuse/adapt architectures created for ImageNet, the accuracy on that primary task could hint at their accuracy on the target task. Indeed, Kornblith \etal~\cite{do_better_imagenet_transfer_better} evaluated 16 CNN architectures primarily created for ImageNet, and transferred for 12 target tasks, and found a strong correlation between primary and target accuracies.

However, as we will see, their findings must be taken with care when applied for skin lesion analysis. In this work, we reproduce their findings for 9 network architectures over the ISIC Challenge 2017 classification task (melanoma vs. all subtasks) --- but find that the correlations disappear when only the top-performing networks are considered.

Creating ensembles of several models are an effective way of both improving accuracies and stabilizing them~\cite{data_depth_design}. The accuracy of machine learning models tends to fluctuate widely due to uncontrolled nuisance factors, like the choice of the training set, and even random conditions, like the initialization of the networks. In this work, we will also evaluate the performance of ensembles in contrast to single models.

The main contributions of this work are:

\begin{itemize}
  \item We evaluate the factors that affect the choice of a CNN architecture for skin lesion analysis. We evaluate 13 factors over 9 architectures on 5 sets of splits created on the ISIC 2017 classification Challenge dataset.
  \item We evaluate the performance of simple ensemble schemes, contrasting them to single-model performances.
\end{itemize}

All our source code is available on GitHub, to allow the community to reproduce our results, from the training of the networks, until the statistical analyses.

We divided this work as follows: In Section~\ref{sec:related_work}, we review the state of the art of transfer learning in skin lesion classification, and discuss current approaches for choosing CNN architectures. We detail the tools, methods, and CNN architectures used in the experiments in Section~\ref{sec:methodology}. We present the results in Section~\ref{sec:results}, and our conclusions in Section~\ref{sec:discussion}.

\section{Related Work}
\label{sec:related_work}

In this section, we briefly review the literature in transfer learning, and model design, in the context of skin lesion classification. We focus on convolutional neural networks, which are the state of the art on the field~\cite{celebi2019dermoscopy}. For more information, the reader may refer to recent works on deep learning for skin lesion analysis~\cite{michel_towards, celebi2019dermoscopy} and for medical images in general~\cite{litjens2017survey}.

The enormous size of deep learning models contrasts with the scarcity of training data for skin lesion analysis, making transfer learning mandatory for that task~\cite{transfer_afonso, data_depth_design}. Indeed, transfer learning is a commonplace procedure for most computer visions tasks. Compared with training from scratch, knowledge transfer not only increases accuracies but also decreases training times~\cite{rethinking_imagenet_pretraining, do_better_imagenet_transfer_better}.

Most CNN architectures were primarily created for ImageNet~\cite{imagenet} --- a very large dataset, with 1.3 million images and 1000 categories. Due to ImageNet's diversity, those networks learn features that generalize well for other tasks~\cite{what_makes_imagnet_good_for_transfer_learning}. That fact was established by the seminal study of Yosinski \etal~\cite{how_transferable}, which quantified the large effect of transfer learning on accuracies.

More recently, Kornblith \etal~\cite{do_better_imagenet_transfer_better} confirmed those findings with an even stronger result, which established a direct correlation between accuracies on ImageNet, and on the target tasks. They evaluated 16 CNN architectures on 12 target tasks and found strong correlations between Top-1 accuracies on ImageNet, and the accuracies on target tasks. The evaluated architectures comprised 5 variations of Inception, 3 of ResNet, 3 of DenseNet, 3 of MobileNet, and 2 of NASNet. The tasks comprised general-purpose computer vision tasks (CIFAR, Caltech, SUN397, Pascal VOC), and more specialized --- but still aimed at natural photos --- tasks (Birdsnap, Food-101, Describable Textures, Oxford Flowers, Oxford Pets, Stanford Cars). They found very strong correlations for both fine-tuned ($R=0.96$, $\rho=0.97$, p-value $<10^{-8}$) and non-fine-tuned ($R=0.99$, $\rho=0.99$, p-value $<10^{-11}$) models. Curiously, the performance of architectures \textit{without transfer} (initialized at random) also showed some correlation ($R=0.37$, $\rho=0.48$, p-value $=0.03$). That shows that the correlation is due not only to the learned weights but also --- although in a lesser degree --- to the architecture design.

None of the Kornblith \etal's tasks are medical tasks. Studies evaluating the importance of transfer learning for medical applications are few, and for skin lesion analysis even fewer. Menegola \etal ~\cite{transfer_afonso} compared several schemes for transfer learning on melanoma classification and found that fine-tuning a network pre-trained on ImageNet gives the better results when compared with using a network pre-trained on another medical task (diabetic retinopathy). Performing a double transfer (ImageNet $\rightarrow$ retinopathy $\rightarrow$ melanoma) did not improve the results, compared with transferring from ImageNet alone. Using an exhaustive factorial design with 7 factors (network architecture, training dataset, input resolution, training augmentation, training length, test augmentation, and transfer learning) over 5 test datasets, for a total of 1280 experiments, Valle \etal ~\cite{data_depth_design} showed that the use of transfer learning is, by far, the most critical factor. In a factorial Analysis of Variance, it explains 14.7\% of the absolute performance variation and 62.8\% of the relative variation (excluding residuals and the choice of test dataset), with high significance (p-value $<0.001$).

In any event, transfer learning and fine-tuning are heavily used for skin lesion classification. All top-ranked submissions for ISIC Challenges 2016~\cite{isic_2016_first_place}, 2017~\cite{isic2017_titans_submission, isic2017_participant_second, isic2017_participant_third, isic2017_participant_first_kerat} and 2018~\cite{isic2018_participants_first_with_data, isic2018_participants_second_with_data, isic2018_participants_first_no_data, isic2018_participants_second_no_data, isic2018_titans_submission} used CNNs pre-trained on ImageNet.

While there is a consensus on the use of transfer learning for skin lesion models, when choosing the architecture no choice is universal. On the contrary, the classification task of the ISIC Challenge shows a progressive \textit{diversification} of architectures. In 2017, the top four submissions used two networks: ResNet~\cite{isic2017_titans_submission, isic2017_participant_second, isic2017_participant_third, isic2017_participant_first_kerat}, and Inception-v4~\cite{isic2017_titans_submission}, while the latest challenge~\cite{isic_challenge_2018}, in 2018, showcased a wider range of choices among the top performers --- not only ResNets and Inceptions, but also DenseNet~\cite{isic2018_titans_submission}, Dual Path Networks~\cite{dual_path_networks}, InceptionResNet~\cite{inceptionv4}, PNASNet~\cite{pnasnet}, SENet~\cite{senet}, among others --- usually in ensembles of multiple architectures~\cite{isic2017_titans_submission, isic2018_participants_first_with_data, isic2018_participants_second_with_data, isic2018_participants_first_no_data, isic2018_titans_submission}.

The best architectures for skin lesion classification remain, thus, an open issue. The Challenge results do not allow analyzing a single factor from a multitude of choices among participants. The study of Valle \etal ~\cite{data_depth_design}, although exhaustive for 9 factors, evaluates only two levels for each factor, picking ResNet-101 and Inception-v4 for architectures. No other study, as far as we know, attempts to answer the question systematically.

In this work, we focus on that issue, by applying the design of Kornblith \etal~\cite{do_better_imagenet_transfer_better} to the task of melanoma classification. We evaluate the performance of 9 networks (listed in Table~\ref{tab:architectures}) pre-trained on ImageNet, and fine-tuned for melanoma classification, on the dataset of the ISIC 2017 challenge.

In comparison to Kornblith \etal's, our study reduces the scope of the tasks and enlarges the scope of the correlations attempted. On the one hand, while they evaluate 12 general and specialized natural-photo tasks, we focus only on melanoma classification. On the other hand, while they correlate only the Top-1 ImageNet accuracy to the target accuracy, we correlate several network attributes and source and target metrics. We aim to obtain hints on how to select the best architectures for melanoma classification.

Since ensembles of models have special importance in the literature of melanoma classification, we also evaluate how simple ensembles of the evaluated models perform in comparison to the models alone.

\section{Methodology}
\label{sec:methodology}

\subsection{Data}

All data comes from the ISIC Challenge 2017 dataset~\cite{isic_challenge_2017}. We do not employ the original training (2000 images), validation (150) and test (600) splits. Instead, we combined all 2750 images and produced five random combinations of train (1750), validation (500), and test (500) splits --- aiming at an approximate 60--20--20\% partition. We chose those proportions because the original validation set was too small to allow making decisions on the deep learning hyper-parameters with confidence, and we made five completely random combinations to allow estimating the variability due to the choice of the splits. The exact images used in our splits are available in our code repository (see next section).

Although the ISIC Challenge 2017 provided annotations for three classes (nevus, seborrheic keratosis, and melanoma), and had three subtasks, we focus only on the subtask of melanoma vs. all.

\subsection{Tools}

We evaluated 9 different architectures (Table~\ref{tab:architectures}), whose PyTorch implementations and pre-trained ImageNet snapshots we obtained from other authors. The choice of publicly available snapshots reflects current practice, since retraining the architectures from scratch on ImageNet is very time-consuming. Kornblith \etal report major performance increases for architectures carefully trained from scratch on ImageNet, for transfer learning \textit{without} fine-tuning. For transfer with fine-tuning (which we evaluate in this paper), they found little difference between public snapshots and models trained from scratch.

Table~\ref{tab:architectures} shows the architectures we chose. Except for MobileNetV2, we chose the architectures for their performance on both ImageNet and the ISIC Challenge. We kept MobileNetV2, present in the original study by Kornblith \etal, since we found interesting to include an architecture aimed at embedded and mobile hardware.

The only modification on the architectures was changing the last layer from the 1000 ImageNet classes to a binary classification layer.

We fine-tuned each network with stochastic gradient descent (SGD) with a starting learning rate of $1\mathrm{e}{-3}$ and momentum factor of $0.9$. All layers were left free to evolve. Whenever the validation loss failed to improve for 8 consecutive epochs, we divided the learning rate by 10. We stopped the tuning if the validation AUC (area under the ROC curve) failed to improve for 16 epochs. To avoid accommodation on the training data order, we shuffled them before each epoch.

We followed the best data augmentation settings proposed in~\cite{dataaug_for_skin}: random horizontal/vertical flips; random cropping; rotation up to $90\degree$; shear up to $20\degree$; area scaling from $0.8$ to $1.2$; and color (saturation, contrast,
brightness, and hue) jittering. We also used augmentation on test with 64 copies, and on validation with 16 copies (average-pooling the probability vectors of the copies as the final result). We resized each image according to the input expected by each architecture ($224\times224$ for DenseNet, ResNet, DualPathNet, SENet, and MobileNetV2; $299\times299$ for Inception-v4, InceptionResNet-v2, and Xception; and $331\times331$ for PNASNet). The inputs were also normalized per-channel using the z-score computed with the training dataset statistics.

We used an NVIDIA Tesla P100 with 12~GiB to fine-tune all models. We considered the memory GPU as a constraint to our experiments, and we picked, for each method the largest batch size (in multiples of 8) that the model could fit (Table~\ref{tab:architectures}). Although in a theoretical setting we could have compared all models with the same batch size, we considered our criterion more useful for practical purposes, since occupying the GPU memory as much as possible is the usual procedure. Considering that practical setting, our criterion is ``fair'' in the sense that it allows considering a compromise between larger models vs. larger batches.

All the source code used in this paper, from model tuning until statistical analyses, is available in our public repository\footnote{\url{https://github.com/learningtitans/cvpr-skin-solo-ensemble}}. Our code is easily adaptable to allow new architectures.

\newcolumntype{Z}{>{\raggedright\arraybackslash}p{0.23\linewidth}}
\newcolumntype{K}{>{\justifying\noindent\arraybackslash}p{0.41\linewidth}}
\newcolumntype{R}{>{\raggedleft\arraybackslash}p{0.052\linewidth}}
\newcolumntype{B}{>{\raggedleft\arraybackslash}p{0.06\linewidth}}

\begin{table*}[!htbp]
\centering
\begin{tabular}{ZRRBRK}
\toprule
\textbf{Architecture} & \textbf{Acc@1} & \textbf{Acc@5} & \textbf{Params (M)} & \textbf{Batch Size} & \textbf{Summary} \\ \midrule
\textbf{DenseNet}~\cite{densenet}~\textsuperscript{A} & 77.6 & 93.8 & 28.7 & 40 & Composed of dense blocks, which concatenate the output feature map of each layer to all subsequent layers. \\
\textbf{Dual Path Nets}~\cite{dual_path_networks}~\textsuperscript{B} & 79.8 & 94.7 & 79.3 & 24 & Combines ResNet's residual paths for feature re-usage and DenseNet's dense connections for new features exploration. \\
\textbf{Inception-v4}~\cite{inceptionv4}~\textsuperscript{B} & 80.2 & 95.2 & 55.8 & 64 & Composed of Inception modules, which have parallel convolutional layers that learn different cross-channel and spatial correlations. \\
\textbf{Inception-ResNet-v2}~\cite{inceptionv4}~\textsuperscript{B} & 80.1 & 94.9 & 42.7 & 32 & Similar to Inception-v4, but with residual connections. \\
\textbf{MobileNetV2}~\cite{mobilenetv2}~\textsuperscript{C} & 71.8 & 91.0 & 3.5 & 128 & Uses depth-wise separable convolutions to produce an efficient network, suitable for mobile devices. \\
\textbf{PNASNet}~\cite{pnasnet}~\textsuperscript{B} & 82.7 & 96.0 & 86.1 & 8 & Designed with modules discovered through Neural Architecture Search (NAS) (current best accuracy on ImageNet). \\
\textbf{ResNet}~\cite{resnet}~\textsuperscript{A} & 78.4 & 94.1 & 60.2 & 56 & Uses residual connections to improve information flow, allowing networks with more than 100 layers. \\
\textbf{SENet}~\cite{senet}~\textsuperscript{B} & 81.3 & 95.5 & 115.1 & 24 & Composed of Squeeze-and-Excitation (SE) blocks, which capture channel-wise dependencies for convolutional features maps. \\
\textbf{Xception}~\cite{DBLP:conf/cvpr/Chollet17}~\textsuperscript{B} & 78.9 & 94.3 & 22.9 & 40 & Extrapolates Inception modules to depth-wise separable convolutions, resulting in more efficient parameter use. \\ \bottomrule
\end{tabular}
\caption{CNN architectures used in the experiment. Acc@1 and Acc@5: performances on ImageNet. Params: number of trainable parameters (in millions). Models and checkpoints sources (superscripts on model names): A) \textit{github.com/pytorch/vision}; B) \textit{github.com/Cadene/pretrained-models.pytorch}; C) \textit{github.com/tonylins/pytorch-mobilenet-v2}.}
\label{tab:architectures}
\end{table*}

\subsection{Experimental Design}
\label{sec:exp_design}

For the main experiment, evaluating the effects on the choice of the architecture, we make 3 repeated experiments (tuning and measurements) for each of the 9 architectures, on each of the 5 sets of splits. The 3 repeated experiments allow evaluating the effect of random choices: initialization of the last layer, dropouts, data augmentation, shuffling of data on epochs, etc. The main experiment, thus, comprises $3\times9\times5=135$ measurements of several metrics: AUC, accuracy, sensitivity, and specificity for both the validation set (at the epoch chosen by the early stopping procedure) and the test set. We also measure the loss at the validation at the epoch chosen.

\begin{figure*}[t]
\centering
\vspace{-0.6cm}
\begin{subfigure}[t]{.63\linewidth}
  \centering
  \includegraphics[width=1.\linewidth]{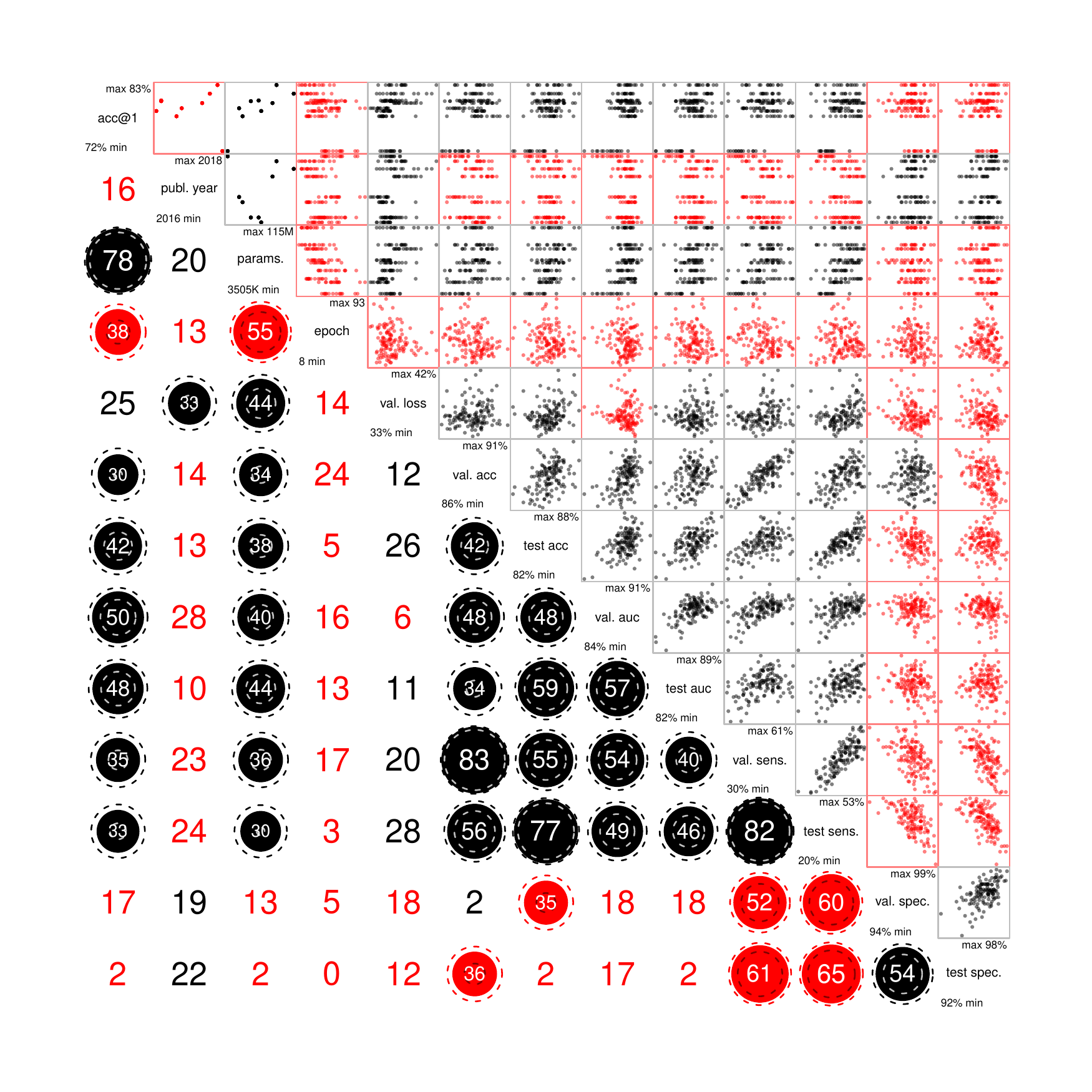}
\end{subfigure}
\\
\begin{subfigure}[t]{.63\linewidth}
  \vspace{-1.6cm}
  \centering
  \includegraphics[width=1.\linewidth]{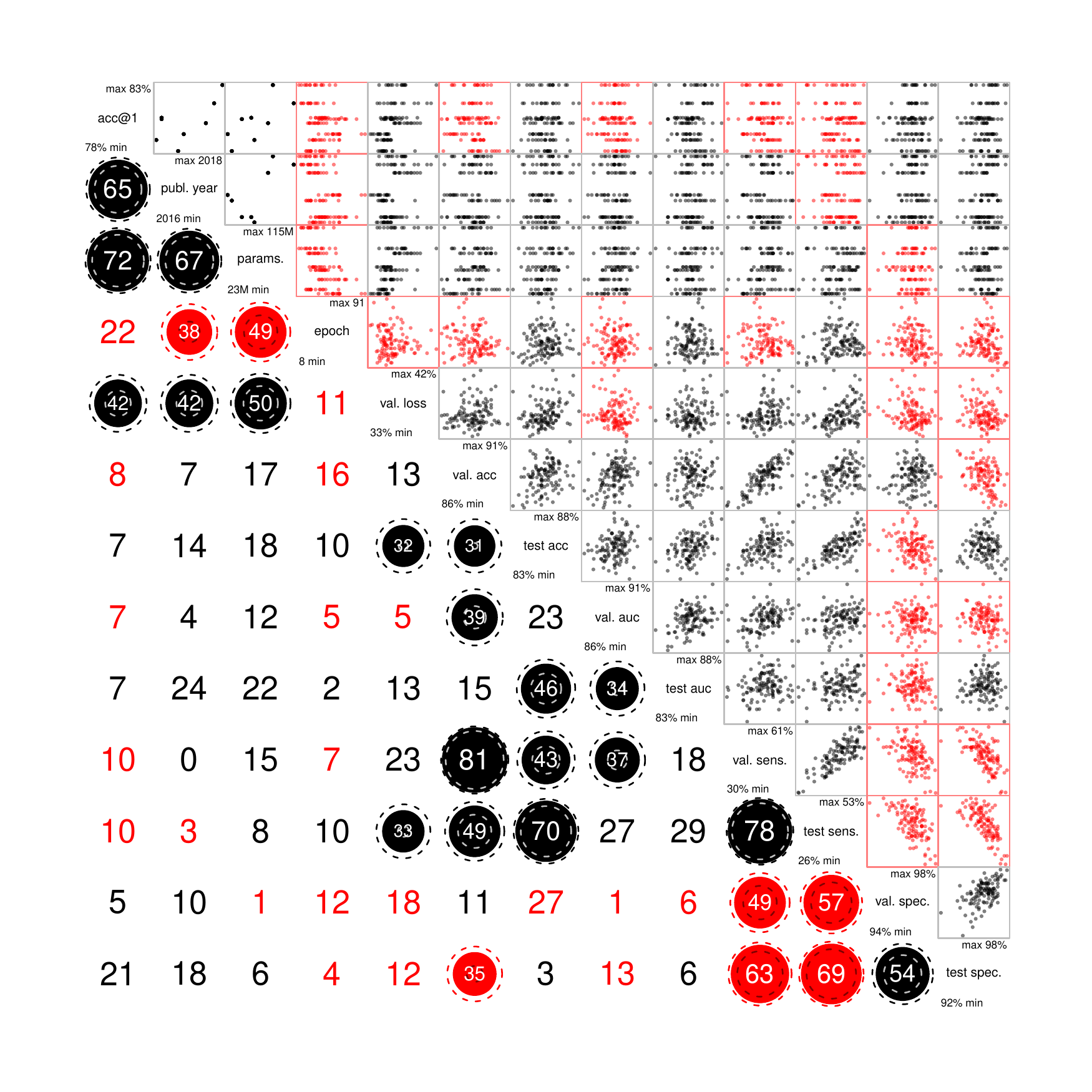}
\end{subfigure}
\vspace{-0.6cm}
\caption{Correlograms of the network attributes and outcome variables. Top correlogram: with MobileNetV2; bottom correlogram: without it. Acc@1: Top-1 accuracy on ImageNet. On each correlogram, the upper matrix has the scatter plots and the lower matrix has the Spearman's $\rho$ correlations (positive in black, negative in red). The area of the circles also indicates the magnitude of the correlations, the dashed circles indicating the confidence interval for $\alpha=0.05/78$ (78 = Bonferroni correction). For intervals containing zero, we omitted the circles, indicating non-significant correlations.}
\label{fig:correlogram}
\end{figure*}

For the analysis of the main experiment, we employ a correlogram (Figure~\ref{fig:correlogram}) of the metrics above, adding 7 attributes of the architectures: Top-1 accuracy on ImageNet, time of publication, number of parameters, and number of the epoch picked by the early stopping. In order to make the correlations comparable across the 5 sets of splits, we perform an adjustment similar to the one used in a repeated measures/within subjects Analysis of Variance: we subtract from each metric on a given experiment its average across all experiments in the same set of splits, and add back its average across all experiments. We consider the correlations significant when their confidence intervals do not contain zero. For the confidence level, we employ a Bonferroni-adjusted  $\alpha=0.05/78$, where 78 is the number of pairs of variables in our correlogram.

For the ensembling experiments, we employ the base models above. For each of the 5 sets of splits, there are 27 single models. We create the ensembles by ordering those 27 models, choosing a number $n$ between 1 and 27, and combining the first $n$\textsuperscript{th} base models into an ensemble. To simplify the evaluation, we use a very simple, but effective~\cite{data_depth_design} strategy of average-pooling the output prediction probabilities for the ensemble.

We contrast two strategies: ordering the base models by their AUC on the validation set and ordering them at random. We replicate the latter 10 times, to evaluate the variability. In all cases, the measurement performed is the AUC on the test set. The experiment aims to determine if the AUC on the validation set is useful to choose the models for the ensemble.

In order to evaluate the results, we employ two plots: in one we contrast the ensembles ordered by the validation set vs. the ensembles ordered at random separately for each of the five sets of splits. In the other, we gather all five splits in a single series for each type of ensemble (validation vs. random) using a repeated measures/within subjects procedure like the one explained above.

\begin{figure*}
\centering
\begin{subfigure}[t]{.82\linewidth}
  \centering
  \includegraphics[width=1.\linewidth]{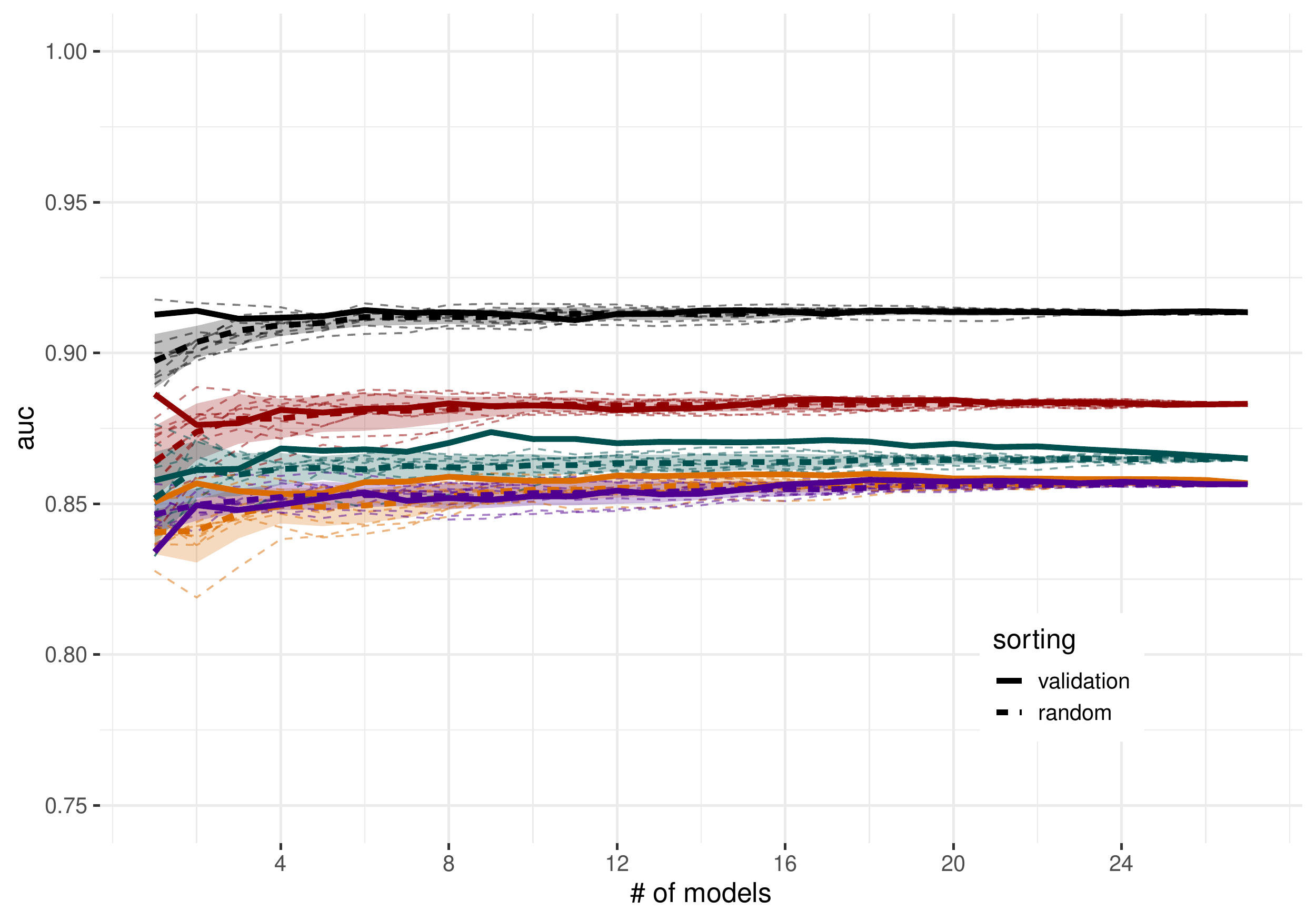}
\end{subfigure}
\\
\begin{subfigure}[t]{.82\linewidth}
  \centering
  \includegraphics[width=1.\linewidth]{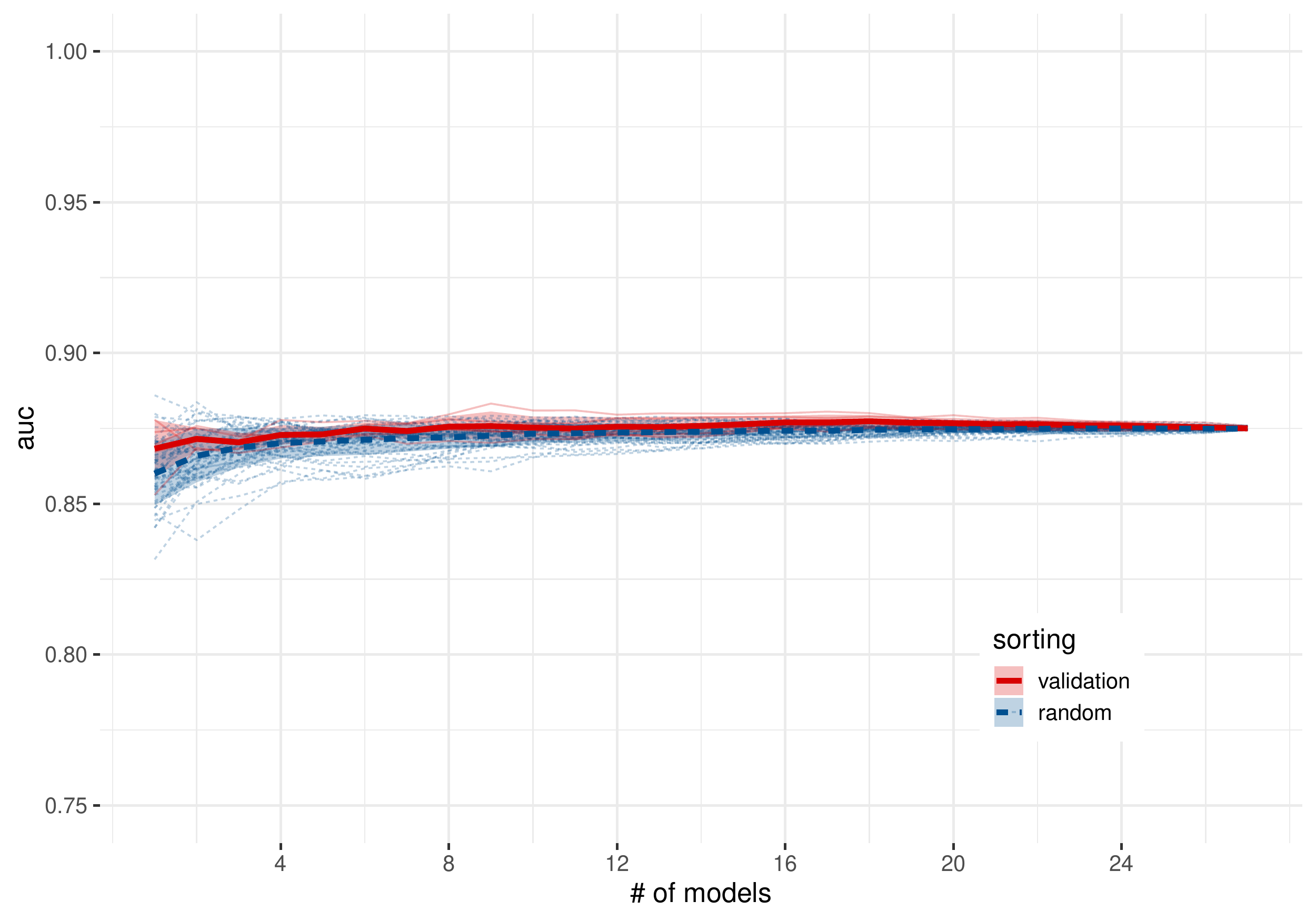}
\end{subfigure}
\caption{Experiments with ensembles of models. Solid lines: models with best validation AUC chosen first. Dashed lines: models chosen at random (Section~\ref{sec:exp_design}). Thick lines: averages, Shadows: standard deviations, Thin lines: individual runs. \textbf{Top}: Experiments separated per split (colors). \textbf{Bottom}: Split differences marginalized, and experiments grouped by type of ensemble. There is a slight advantage in using the validation set to choose the models, but choosing at random also provides good results.}
\label{fig:ensembles}
\end{figure*}

\section{Results}
\label{sec:results}

The correlation analyses are in Figure~\ref{fig:correlogram}, the results on the topmost correlogram appear to partially confirm the results of Kornblith \etal. Indeed, the first column show positive significant Spearman's correlations ($\rho$) between the Top-1 ImageNet accuracy with almost all the target measure metrics: accuracy, AUC, sensitivity, and specificity. However, a more careful inspection of the data --- on the scatter plots of the first line --- raises questions about that conclusion. The positive correlations seem to be due to a single group of samples having detached from the rest of the data. In addition, while Kornblith \etal. observed very high correlations ($\rho>0.95$), the ones we found, although significant, are modest, to say the least ($0.33<\rho<0.50$). The number of parameters in the network also show significant modest correlations with most metrics ($0.30<\rho<0.44$).

The correlogram on the bottom of Figure~\ref{fig:correlogram} further contradicts the findings of Kornblith \etal. By removing MobileNetV2 from the analysis, all significant correlations between ImageNet accuracy and target metrics disappear. Not even a general tendency appears: half the remaining (non-significant) correlations are positive, and the other half negative. The correlations with the number of parameters also become non-significant, although a general tendency still remains: the non-significant correlations remain mostly positive.

On both correlograms, we observe a general positive correlation between the metrics, clearer on the top correlogram (with MobileNetV2). The exception is specificity, which not only shows the expected anti-correlation with sensitivity, but also tends to correlate negatively with most other metrics. Those tendencies had been already observed by Valle \etal~\cite{data_depth_design} on the evaluation of two architectures.

The most useful significant result in both correlograms is the positive correlation between the metrics on the validation set and on the test set --- especially when contrasted with the non-significant or much smaller correlations between the validation loss and the metrics. That suggests that the validation loss might be a poor proxy for \textit{any} of the metrics, and that using the actual metric to make decisions (e.g., on early stopping) might be a better plan.

Two results on the correlograms reinforce the importance of ensembles for skin lesion classification: first, the impossibility of establishing any definitive criterion to select an architecture as a definitive choice. Second, the large amount of variability due to uncontrollable sources (\ie choice of the folds and random nuisances).

The results of the experiments on ensembles are on Figure~\ref{fig:ensembles}. In the topmost plot, the results are separated for each set of splits (each represented by a different color) --- the most influential factor in determining the results. For each set, however, we can see how choosing models ordered by the validation split tends to produce better ensembles.

The bottom plot marginalizes over the splits for both types of ensembles, and shows how the ensembles sorted by validation have a slightly better mean (thick lines) and smaller variability (shaded area). It is remarkable, however, how even ensembles of enough models chosen at random have surprisingly good performance.

In both plots, we can see how variability decreases as the ensembles incorporate more models. Those estimations, however, must be interpreted carefully: the variability decreases because there is a limited number of available base models, and thus the ensembles necessarily become more alike as their number of base models increase --- at 27 base models, all ensembles become the same, and any variability disappears.

\section{Discussion}
\label{sec:discussion}

We were disappointed our results failed to confirm those found by Kornblith \etal\ --- had we found the same results as them, we could employ the accuracy on ImageNet as a safe proxy to choose architectures for skin lesion analysis. In our results, however, not only no network characteristic (e.g., number of parameters) correlates well with the target metrics, but also most networks appear impossible to distinguish from one another in an statistical test. Uncontrollable factors such as the choice of splits on the dataset, and even random nuisances appear more influential than the choice of architectures. That analysis however, is only true for a selection of very high-performance architectures. When we add a shallower, less complex architecture (MobileNetV2) to the lot, it appears different than all others --- to the point that just by creating a set of very different measurements, it can make several of the correlations significant. Our results certainly do not imply that one can select an architecture from 2013, and expect 2019-level performances.

It is not obvious why the findings of Kornblith \etal do not reappear in our study. On one hand, our study has an important limitation: the size of the dataset. All datasets employed by them were larger than ours (although one of them had a training set even smaller than ours). In a follow-up study, we would like to do our correlations over several datasets, including the full ISIC Archive. On the other hand, \textit{their} study has also an important limitation: they did not evaluate several sets of splits for each dataset, neither multiple trainings for each network. We found those uncontrolled factors to be the main sources of variability, largely reducing the correlations. Finally, an important distinction between both studies, is that they perform extensive tuning of the training hyper-parameters of their models, while we adopt a standard approach considered ``best practice'' for most models. We think those choices do not necessarily reflect a limitation of either study, but different aims. Instead of extensive tuning to a particular skin lesion dataset, we opt for several attempts, to reflect the variability expected on real-world scenarios. They, however, are evaluating ``classical'' computer vision tasks, where those extensive tunings are expected to reflect models present in existing literature.

Those results reinforce the importance of ensembles of diverse architecture as the preferred mechanism to obtain good models for skin lesion analysis.  Our results show that for small ensembles it is very useful to employ the validation set to select the best base models, but that for large ensembles one can possibly get away simply choosing the models at random.

Although the aim of this paper was not to maximize any of the measured metrics, the plots on both Figures~\ref{fig:correlogram} and~\ref{fig:ensembles} help as sanity checks, to verify that our models' performances are not unrealistically low compared to existing art. The melanoma vs. all AUCs of the single models evaluated was between 84 and 91\% (86 and 91\% without MobileNetV2). The first place on the ISIC 2017 Challenge was 87.4\% --- almost exactly the average value we found for our ensembles.

\section*{Acknowledgments}

S. Avila is partially funded by Google LARA 2018. E. Valle is partially funded by a CNPq PQ-2 grant (311905/2017-0). This work was funded by grants from CNPq (424958/2016-3), FAPESP (2017/16246-0) and FAEPEX (3125/17). The RECOD Lab receives addition funds from FAPESP, CNPq, and CAPES. We gratefully acknowledge NVIDIA for the donation of GPU hardware.

{\small
\bibliographystyle{abbrv}
\bibliography{egbib}
}

\end{document}